\documentclass[a4paper,12pt,journal]{IEEEtran}
\usepackage[utf8]{inputenc}
\usepackage[T1]{fontenc}
\usepackage{fixltx2e}
\usepackage{graphicx}
\usepackage{grffile}
\usepackage{longtable}
\usepackage{wrapfig}
\usepackage{rotating}
\usepackage[normalem]{ulem}
\usepackage{amsmath}
\usepackage{textcomp}
\usepackage{amssymb}
\usepackage{capt-of}
\usepackage{hyperref}
\usepackage{url}
\usepackage{subfigure}
\usepackage[font=small,skip=2pt]{caption}
\usepackage{authblk}
\usepackage{float}
\usepackage{booktabs}
\usepackage{color}
\usepackage{hyperref}
\hypersetup{colorlinks,breaklinks,linkcolor=blue,urlcolor=blue,anchorcolor=blue,citecolor=blue}
\author[]{Eduardo Castelló Ferrer}
\affil[]{MIT Media Lab, 75 Amherst St., Cambridge, MA. \\ ecstll@mit.edu}
\author{Eduardo Castelló Ferrer}
\date{}
\title{A wearable general-purpose solution for Human-Swarm Interaction}
\hypersetup{
 pdfauthor={Eduardo Castelló Ferrer},
 pdftitle={A wearable general-purpose solution for Human-Swarm Interaction},
 pdfkeywords={},
 pdfsubject={},
 pdfcreator={Emacs 25.1.1 (Org mode 8.3.6)}, 
 pdflang={English}}
\begin{document}

\maketitle
\begin{abstract}
Swarms of robots will revolutionize many industrial applications, from
targeted material delivery to precision farming. Controlling the
motion and behavior of these swarms presents unique challenges for
human operators, who cannot yet effectively convey their high-level
intentions to a group of robots in application. This work proposes a
new human-swarm interface based on novel wearable gesture-control and
haptic-feedback devices. This work seeks to combine a wearable gesture
recognition device that can detect high-level intentions, a portable
device that can detect Cartesian information and finger movements, and
a wearable advanced haptic device that can provide real-time feedback.
This project is the first to envisage a wearable Human-Swarm
Interaction (HSI) interface that separates the input and feedback
components of the classical control loop (input, output, feedback), as
well as being the first of its kind suitable for both indoor and
outdoor environments.
\end{abstract}

\section{Human-Swarm Interaction: the emergent field}
\label{sec:orgheadline3}

\begin{figure}[htb]
\centering
\includegraphics[width=\columnwidth]{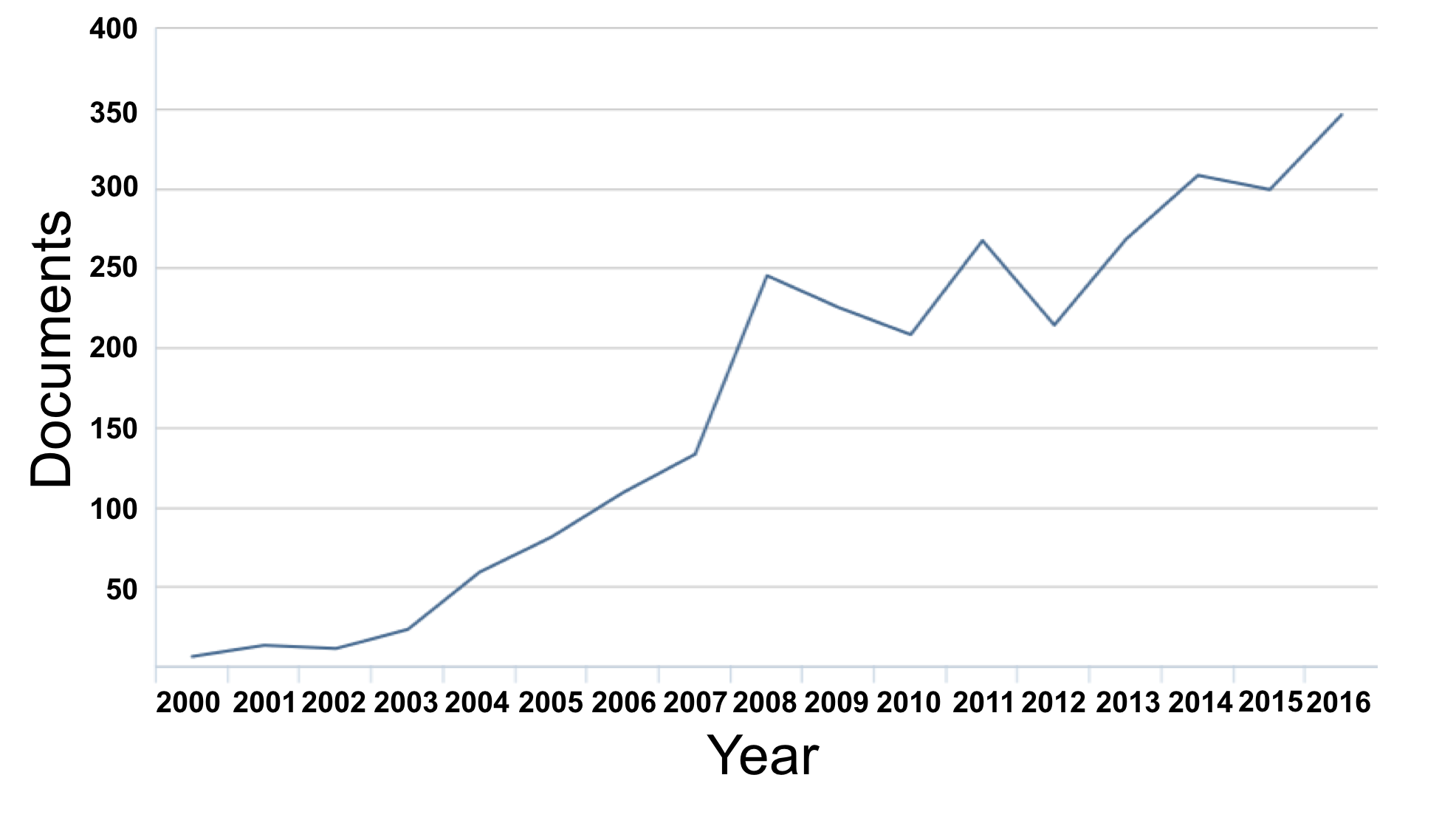}
\caption{\label{fig:orgparagraph1}
Total number of swarm robotics systems research documents published annually from 2000 to 2016.}
\end{figure}

With a strong initial influence from nature and bio-inspired models
\cite{Walker2011,Bonabeau1999}, swarm systems are known for
their adaptability to different environments \cite{Bentes2012} and
tasks \cite{Brambilla2013}. As a result, swarm robotics research has
recently been gaining popularity –- Fig.
\ref{fig:orgparagraph1}\footnote{Information retrieved from \href{http://www.scopus.com}{Scopus} research database.} –-. As the cost of robotic
platforms continues to decrease, the number of applications
involving multiple robots is increasing. These include targeted
material transportation \cite{Chen2013}, where groups of small robots
are used to carry tall, and potentially heavy, objects; precision
farming \cite{Emmi2014,SwarmFarm}, where a fleet of autonomous agents
shifts operator activities in agricultural tasks; and even
entertainment systems \cite{Alonso-Mora2014}, where multiple robots
come together to form interactive displays.

The efficiency of performing tasks with robotic teams depends on two
main factors: the level of robot autonomy, and the ability of human
operators to command and control the team of robots. Regarding the
latter, the transition from current application scenarios where
several human operators control a single robot \cite{Yanco2015} to
environments where a single human control multiple robots, has been
identified as one of the main challenges in robotics research
\cite{Hayes2014,Brambilla2013,Kolling2015}.

One of the clearest examples of this necessity is when the task
conducted by the team of robots becomes extremely complex and
begins to require high-level, cognitive-based decisions inline
(e.g., exploration of dynamic, unstructured, and unpredictable
environments for search and rescue applications). When a robot swarm
needs to react to or quickly respond to an abrupt event (e.g., a
fast stop), the absence of human intervention can even lead to
complete mission failure. In these situations, full autonomy is
still far from being reached by robot units alone, and human
intervention is necessary for adequate performance. 

\begin{figure}[htb]
\centering
\includegraphics[width=\columnwidth]{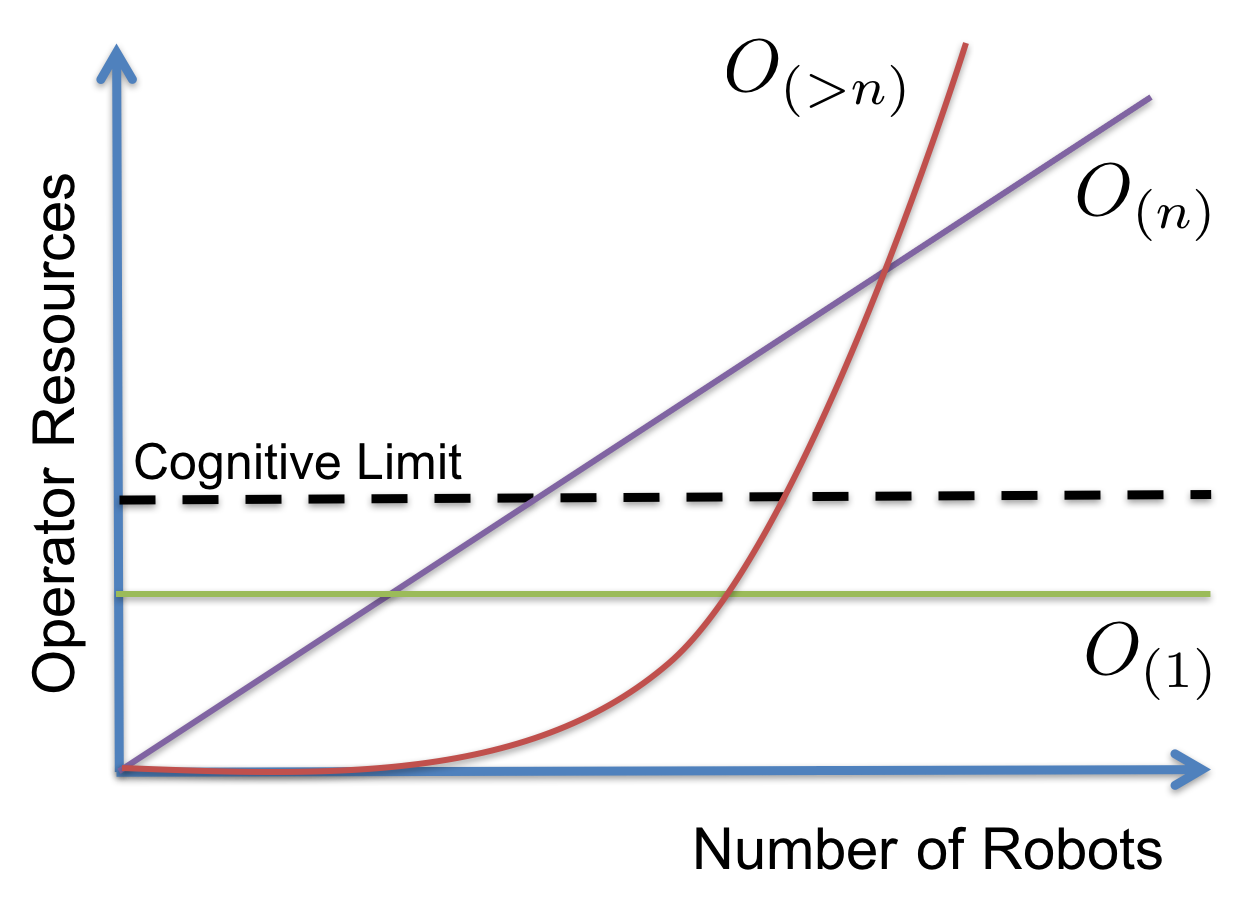}
\caption{\label{fig:orgparagraph2}
Illustration of the different complexity levels and their progressions in the field of human-swarm interaction.}
\end{figure}

However, the ability to command a swarm of robots requires a
significant cognitive effort from human operators. Previous works
\cite{Kolling2015,Lewis2013}, have emphasized the complexity
of these tasks and have compared them to computational complexity
(\(O\)). Likewise, swarm operators traditionally perform a repetitive
sequence of steps to enable the system (i.e., the robot swarm) to
fulfill an objective, or reach some desired goal state
\cite{Chien2012}. Normally, these sequences of steps become more
complex as the operator has to share his/her cognitive resources
among a higher number of robots \cite{Crandall2005}.

Under this framework, different command and control operations
involving robot swarms can have different levels of complexity (Fig.
\ref{fig:orgparagraph2}). For instance, control modes, such as the
leader-follower approach \cite{Setter2015-leaderfollower} where the
number of possible actions (\(n\)) is independent of the number of
robots, can represent a relatively low-level of complexity (ideally
\(O_{(1)}\)) for human operators under their cognitive limit. In
contrast, if several robots are performing independent tasks, the
complexity level might increase linearly as new robots and tasks are
included into the swarm (\(O_{(n)}\)), eventually surpassing the
cognitive abilities of the operator and making the operation of the
swarm unsustainable. Moreover, task scenarios where robots need to
tightly coordinate (e.g., transporting objects with deformable
\cite{Alonso-Mora2015-deformable} shapes) are considered to have an
exponential complexity level (\(O_{(>n)}\)) due to the
inter-dependencies between robots, making the operation of such
group of robots even harder.

The primary purpose of this cognitive complexity framework was to
emphasize the effort of human operators required to control a swarm
robotics system, and the basic need of creating tools and techniques
that allow operators to control higher number of robots without
reaching their cognitive limits.

\begin{figure}[htb]
\centering
\includegraphics[width=\columnwidth]{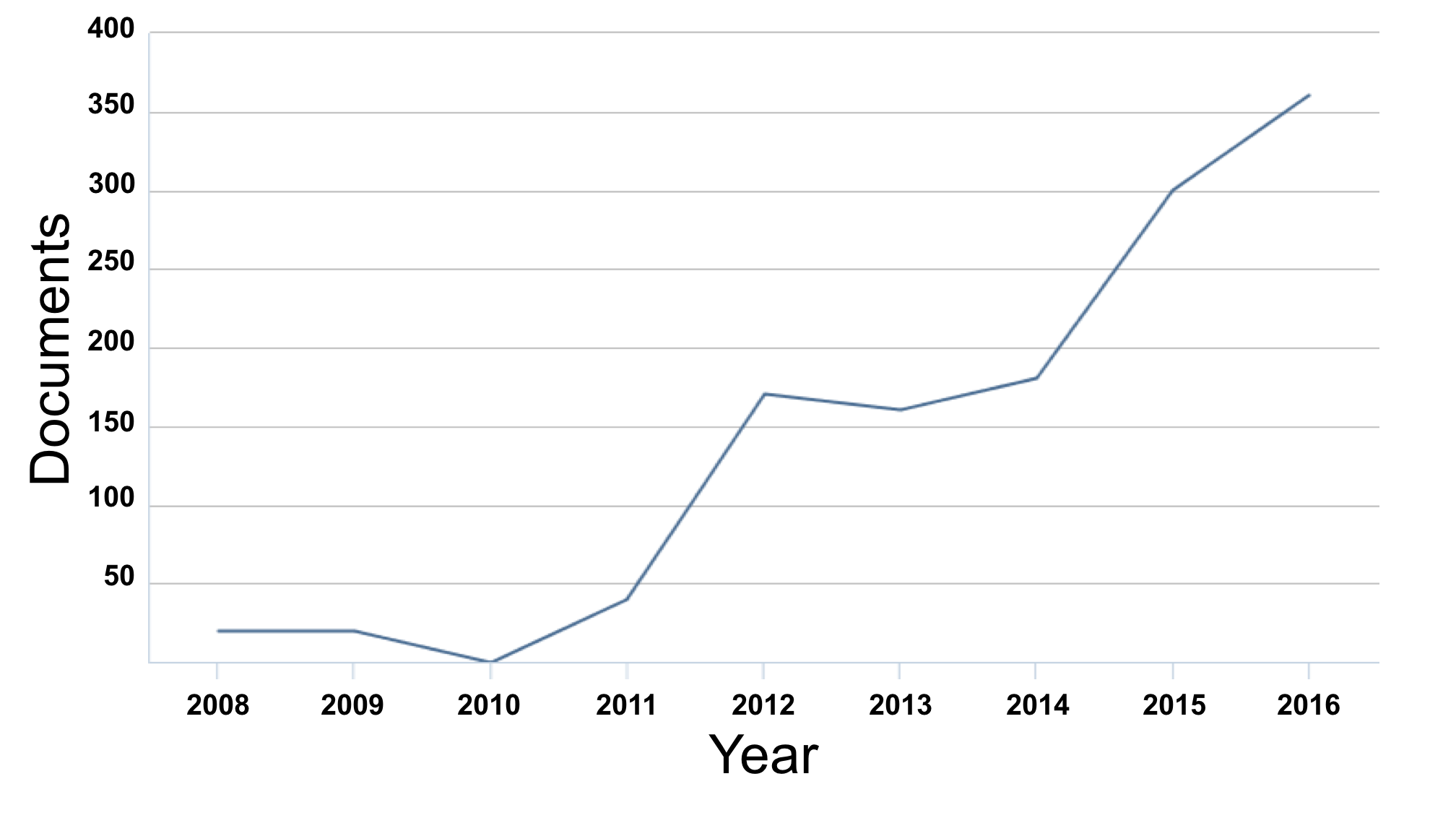}
\caption{\label{fig:orgparagraph3}
Total number of human-swarm interaction research documents published annually from 2008 to 2016.}
\end{figure}

Human-Swarm Interaction (HSI) is a prominent research field –- Fig.
\ref{fig:orgparagraph3}\footnotemark[1]{} –- that aims to allow a human
operator to be aware of certain swarm-level information that he/she
can use to make decisions regarding the swarm behavior. However,
this is a complicated process since some kind of mechanism is needed
to bridge the information gap between the human operator and the
robot swarm. Human supervision normally relies on global goals such
as mission statements or route planning. In contrast, simple robot
units are usually hardware-limited and can access only to local
information.

The design of interfaces that allow operators to control a swarm of
robots is receiving increasing research attention
\cite{Nunnally2013,Stoica2013,Haas2011}. Several well-known
technologies -- including vision-based systems, haptic devices and
electromyographical (EMG) receptors -- have been proposed. However,
seamless interaction between operators and robot swarms has not yet
been achieved, not only due to the complexities of translating
numerous local information streams (i.e., the robot swarm) to a
unified global input scheme (i.e., the human operator), but also due
to the complex infrastructure settings of existing interfaces such
as vision-based sensors or global positioning systems, which only
work in controlled environments, and a lack of appropriate feedback
that can guide the operator and provide accurate information about
the swarm’s state. These obstacles notwithstanding, a
general-purpose human-swarm interface is required to tackle the next
wave of challenges facing industry and advance the technology to a
new state of the art.

In the following, I will discuss two promising technologies that, if
combined, could support the development of a general purpose HSI
interface to control robotic swarms in an efficient and natural
manner. 

\subsection{Gesture Recognition: a versatile high-level input mechanism}
\label{sec:orgheadline1}

Gestures and body movements are a natural way to communicate
intentions and strengthen messages. Gestures are part of our social
communication skill set \cite{Mundy2003}, which humans can use,
understand and analyze. Hand gestures were very early adopted in
research on human-robot interaction \cite{Torige1992}. However, it
took more than 20 years to utilize them in HSI to convey an
operator’s intentions to small swarms of robots \cite{Nagi2012}.

\begin{figure}[htb]
\centering
\includegraphics[width=\columnwidth]{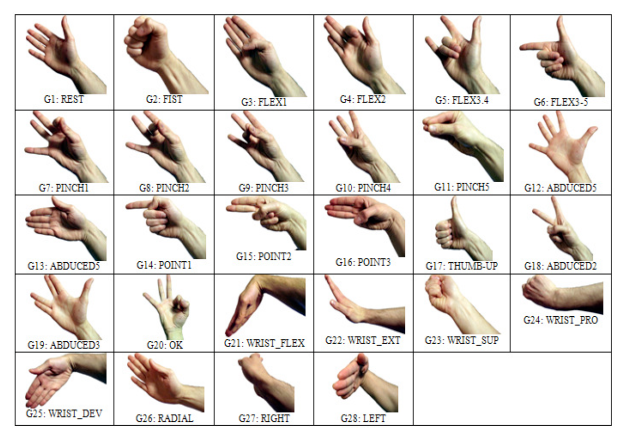}
\caption{\label{fig:orgparagraph4}
Gesture taxonomy proposed in \cite{Stoica2013}.}
\end{figure}

\begin{figure}[htb]
\centering
\includegraphics[width=\columnwidth]{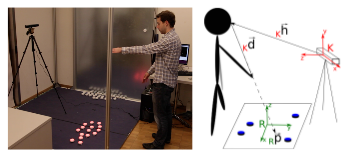}
\caption{\label{fig:orgparagraph5}
HSI scenario proposed in \cite{Alonso-Mora2015}. This configuration is intended for remote interaction applications for which a vision-based sensor (Kinect) is used to extract the body posture of the operator, which is communicated to the robot swarm through a centralized controller. A global positioning system (Vicon) is used to guide robots to their target positions.}
\end{figure}

\begin{figure}[htb]
\centering
\includegraphics[width=\columnwidth]{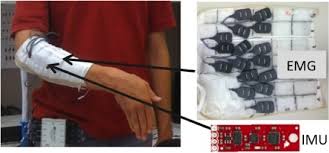}
\caption{\label{fig:orgparagraph6}
BioSleeve HSI device developed by NASA JPL \cite{Stoica2013}.}
\end{figure}

Recently, gesture-based HSI has evolved with the development of
rich gesture taxonomies –- e.g., Fig. \ref{fig:orgparagraph4}
–-, which operators employ to control a group of robots. These
taxonomies have mainly focused on remote interaction (i.e.,
tele-operation) applications conducted in controlled indoor
environments \cite{Alonso-Mora2015,Stoica2013}. Despite the high
correct classification rates (CCRs) and solid conceptual foundation
for future research achieved by these works, their models are
difficult to use in other experimental settings as they require
complex infrastructure such as vision-based sensors and global
positioning systems –- Fig. \ref{fig:orgparagraph5} –- or specialized hardware
–- Fig. \ref{fig:orgparagraph6} –-.

\begin{figure}[htb]
\centering
\includegraphics[width=\columnwidth]{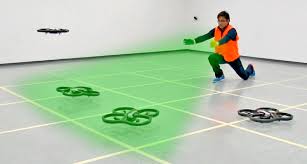}
\caption{\label{fig:orgparagraph7}
Proximity HSI. Robots require a line of sight in order to recognize operator's gestures \cite{Nagi2015}.}
\end{figure}

A different approach was proposed in \cite{Nagi2015}, in which robots
had to distinguish in a distributed fashion the orders and commands
provided by the operator. This method was designed to enable the
operator to interact with the swarm in a proximity environment –-
Fig. \ref{fig:orgparagraph7} –-, making the human operator a ‘special’ swarm
member. However, the robots required a direct line of sight to the
operator in order to detect and classify the operator’s gestures. A
consensus mechanism was then used within the robot swarm to reach
an agreement about the operator’s intentions. Its lack of complex
infrastructure makes this method suitable for a wider range of
scenarios, such as on-the-spot progress checks of a swarm’s
operation, as well as in outdoor environments. However, the
approach is nonetheless hamstrung by the limited sensing and
computational power of individual robots, and so it is unclear if
or how it could be applied to large swarms \cite{Kolling2015}.

Even though the above-described remote and proximity interaction
approaches are promising steps towards achieving suitable
gesture-based control methods for specific applications, they have
key limitations. Methods proposed for remote interaction rely on
complex infrastructure, while proximity interaction methods suffer
from scalability issues. Despite these problems, the aforementioned
works prove that gestures can be a feasible way to control a swarm
of robots in both remote and proximity interaction scenarios. Given
enough flexibility, they may be used in a more general interface
that could be suitable in both application scenarios in the future. 

\subsection{Haptic Feedback: augmented assistance for the operator}
\label{sec:orgheadline2}

Another popular approach to combine robot swarms with human input
has been to explore the haptic channel. Haptic technology provides
a way in which information related to swarm status can be
transferred back to the operator via tactile or force feedback. 

\begin{figure}[htb]
\centering
\includegraphics[width=\columnwidth]{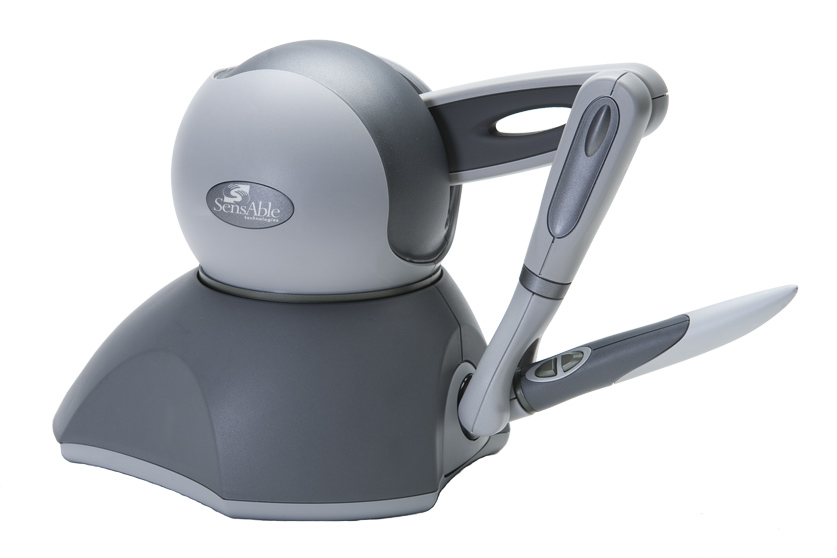}
\caption{\label{fig:orgparagraph8}
Phantom OMNI Haptic Device by SensAble Technologies.}
\end{figure}

\begin{figure}[htb]
\centering
\includegraphics[width=\columnwidth]{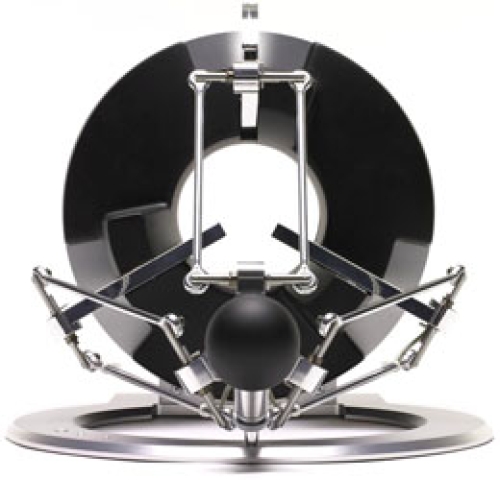}
\caption{\label{fig:orgparagraph9}
Omega 3 Haptic Device by Force Dimension.}
\end{figure}

Haptic devices such as \href{http://www.dentsable.com/haptic-phantom-omni.htm}{Phantom Omni} -- Fig. \ref{fig:orgparagraph8} -- or
\href{http://www.forcedimension.com/downloads/specs/specsheet-omega.3.pdf}{Omega3} -- Fig. \ref{fig:orgparagraph9} -- have been extensively used to
orchestrate the movements of whole swarms of robots
\cite{Nunnally2013SAGE}, certain subgroups of the swarm
\cite{Nunnally2013} or robot teams using a leader-follower approach
\cite{Setter2015}.

In previous research, haptic feedback has been used in combination
with existing methods such as continuous visual input to assist a
human operator \cite{Haas2011}. Haptic information has proven useful
in guiding the operator in situations where a robot swarm is
operating in obstacle-populated environments \cite{Nunnally2013} or
unstructured areas \cite{Hong2013}. In such scenarios,
attraction/repulsion forces are calculated according to
environmental obstacles or swarm members’ positions, and
transferred back to the operator to assist his/her decisions.

Even though current haptic devices provide a way to receive
feedback, they suffer from several limitations. Such devices rely
only on Cartesian input information (trajectories, vectors, etc.)
but cannot process high-level commands such as those provided using
gesture-based interaction. Further, they unify input and feedback
components, which could confuse the operator in situations where
input and elaborate haptic feedback signals occur simultaneously.
For instance, the operator might need feedback about the energy
status of a robot at the same time as he/she is trying to
accurately guide it through an obstacle-free path. 

Another drawback of devices such as the Omega3 and Phantom Omni is
that they represent a single point of failure in case of
malfunction, which might be a problem in terms of building a
general-purpose interface. Finally, one of their most significant
constraints is that they are limited to indoor environments
equipped with a computer terminal, thus excluding missions in
outdoor or other environments.  

\section{A wearable gesture-haptic interface for human-swarm interaction}
\label{sec:orgheadline8}

The gaming and wearable technology industries have been a good
source of breakthroughs and disruptive devices not only for
commerce, but for academic research as well. Devices such as the
\href{http://dev.windows.com/en-us/kinect}{Microsoft Kinect} or \href{http://www.oculus.com/en-us}{Oculus Rift}, initially designed as interactive
controllers for common console platforms and game engines, have been
extensively used within the robotics community to assist key
research activities. In the last few years, we have observed how the
second generation of wearable and gaming devices has reached the
market with a special focus on haptic and monitoring capabilities.
As the cost of these new devices decreases and their technology
starts to provide enhanced features, novel application domains open
up for their use in robotics research. 

\begin{figure}[htb]
\centering
\includegraphics[width=\columnwidth]{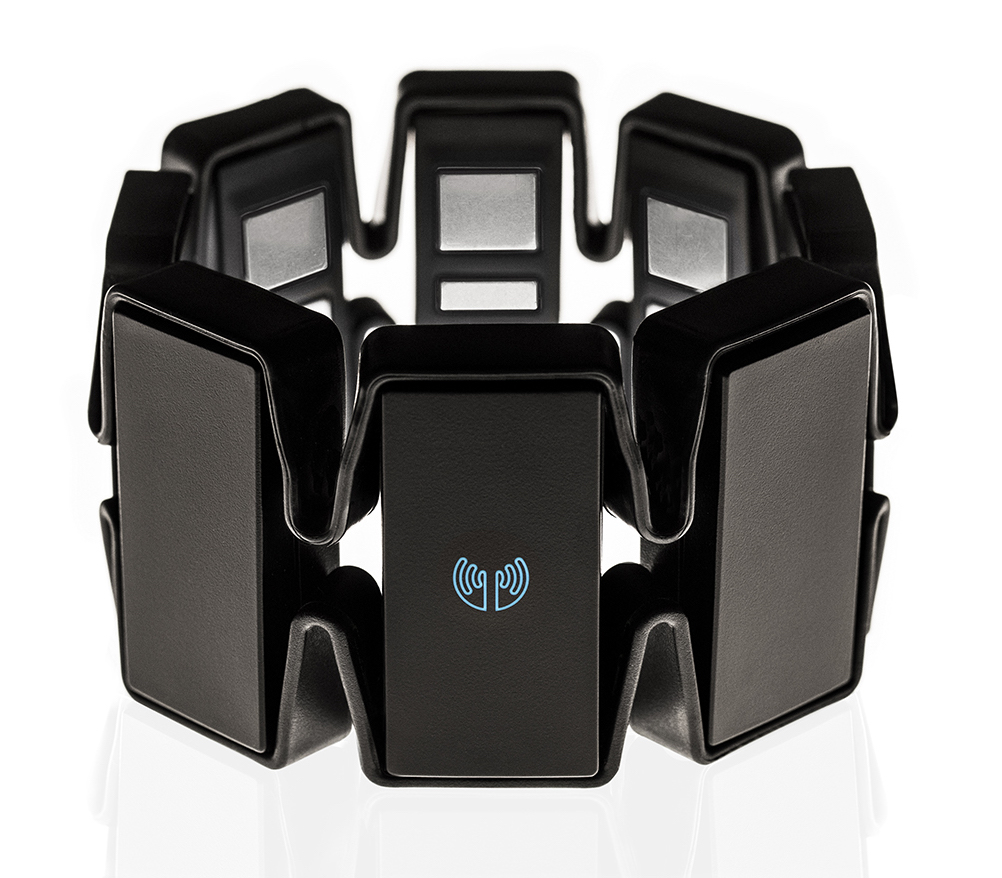}
\caption{\label{fig:orgparagraph10}
Myo armband by Thalmic Labs.}
\end{figure}

\begin{figure}[htb]
\centering
\includegraphics[width=\columnwidth]{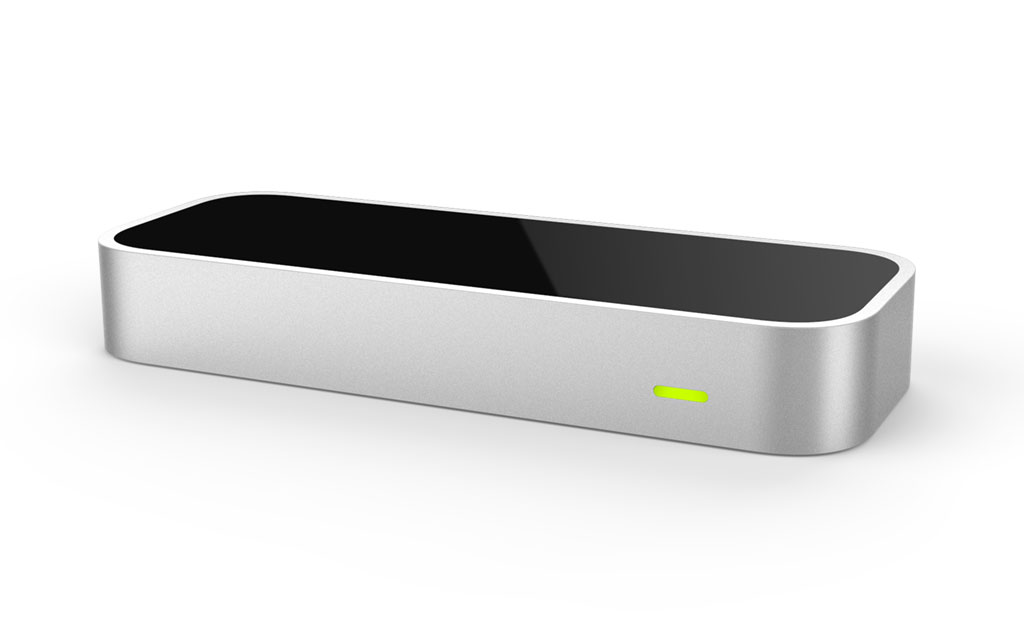}
\caption{\label{fig:orgparagraph11}
Leap motion by Leap Motion Inc.}
\end{figure}

Gesture recognition devices such as the \href{http://www.thalmic.com/en/myo/}{Myo armband} -- Fig. \ref{fig:orgparagraph10}
-- or \href{http://www.leapmotion.com}{Leap motion} -- Fig. \ref{fig:orgparagraph11} -- demonstrate the
feasibility of transferring gestures and simple commands to a
computer system. Their low cost and suitability to both indoor and
outdoor environments make them interesting candidates for the
gesture recognition component of a general-purpose interface.

\begin{figure}[htb]
\centering
\includegraphics[width=\columnwidth]{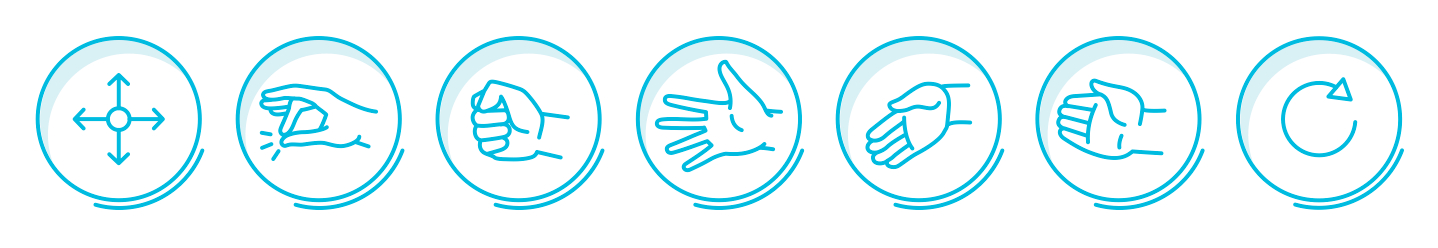}
\caption{\label{fig:orgparagraph12}
List of gestures and motion patterns recognized by Myo.}
\end{figure}

Myo is a wearable armband device that can recognize a rich set of
gestures –- Fig. \ref{fig:orgparagraph12} -– by using 8 EMG muscle sensors
installed in its frame. Its built-in accelerometers and gyroscopes
also allow Myo to detect arm motion accurately. Several vibration
motors give the additional possibility of providing haptic feedback
in the form of short, medium and long vibrations. Myo is equipped
with a rechargeable lithium-ion battery which is designed to last a
day of continuous use. Finally, its Bluetooth connectivity allows
Myo to transmit data to a computer or external device.

\begin{figure}[htb]
\centering
\includegraphics[width=\columnwidth]{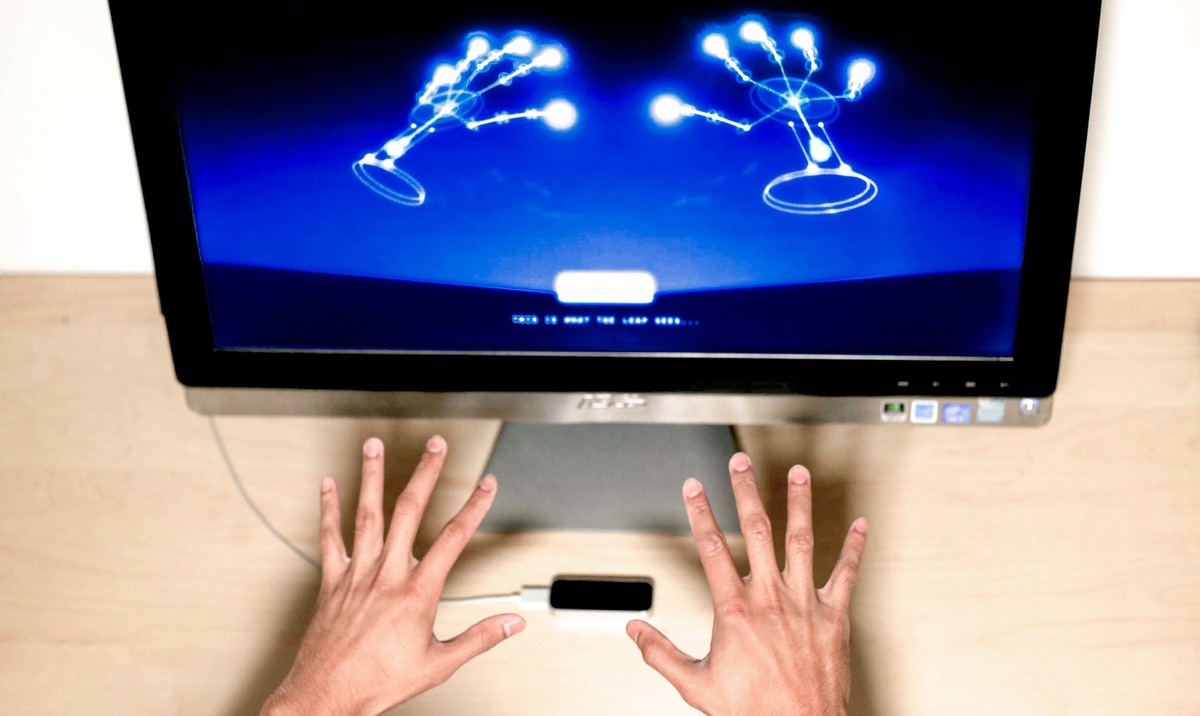}
\caption{\label{fig:orgparagraph13}
Leap motion in action.}
\end{figure}

Leap Motion is a portable 3D motion sensor that is able to detect
hand movements as well as finger positions in a large interaction
space (eight cubic feet) -- Fig. \ref{fig:orgparagraph13} --. The core
of the device consists of two cameras that track the light reflected
by three built-in infrared LEDs. Leap Motion has been used in
academic research focused on fingers and their movements, such as on
sign language recognition \cite{Ching-Hua2014}.

Leap Motion is a great complement to Myo since they each detect
different types of information. Myo can recognize high-level
intentions (gestures) whereas Leap Motion can recognize Cartesian
information such as paths or routes, as well as deictic information
like numbers through finger patterns. The possibility of combining
these types of information allows for the creation of a wide range
of control modes. For instance, high-level gestures could trigger
‘actions’ such as ``follow path'' or ``change parameter value'', while
deictic patterns could provide detailed information about the shape
of the path to follow or the value of the parameter to change. 

\begin{figure}[htb]
\centering
\includegraphics[width=\columnwidth]{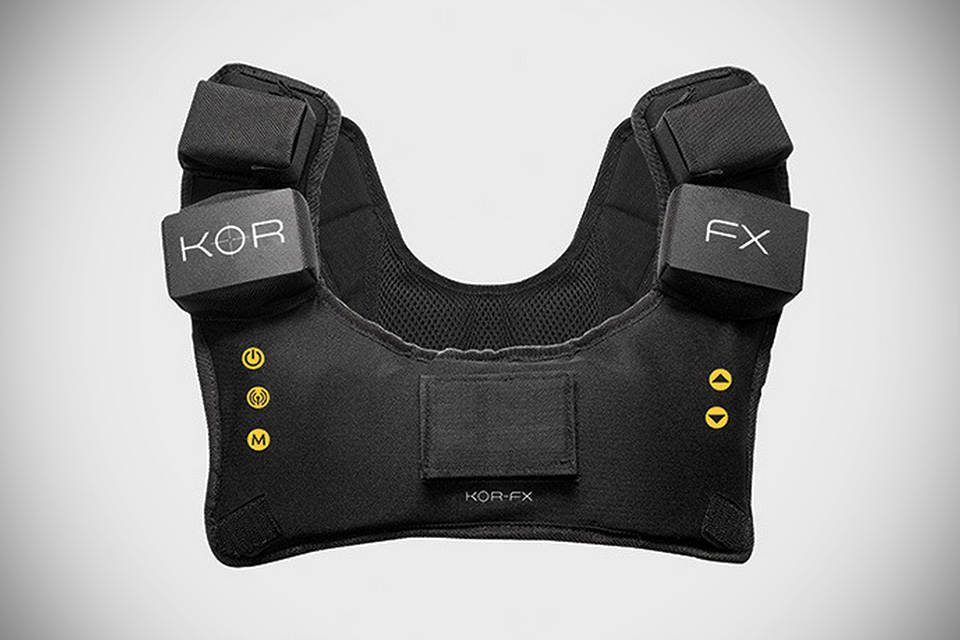}
\caption{\label{fig:orgparagraph14}
KOR-FX Haptic Suit by Immerz, Inc.}
\end{figure}

\begin{figure}[htb]
\centering
\includegraphics[width=\columnwidth]{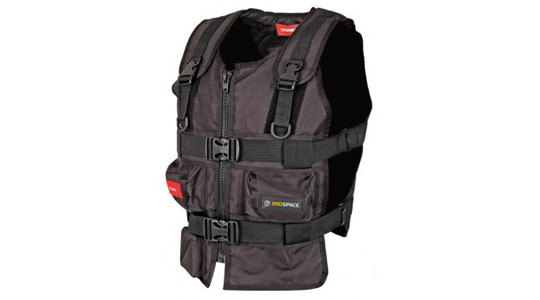}
\caption{\label{fig:orgparagraph15}
3\(^{\text{rd}}\) Space Suit by TN Games.}
\end{figure}

Advanced haptic devices have been recently introduced into the
gaming industry to increase the immersion experience while playing
video games. Gaming vests such as \href{http://www.korfx.com/start}{KOR-FX} -- Fig. \ref{fig:orgparagraph14}
-- or \href{http://tngames.com/}{3\(^{\text{rd}}\) Space} -- Fig. \ref{fig:orgparagraph15} -- allow video
game players to deeply engage with a game’s current events by
producing vibrations (KOR-FX) or pneumatic pulses (3\(^{\text{rd}}\) Space)
according to the in-game situation. Users of these devices obtain
rich information such as direction, intensity, and rhythm.
Transforming swarm-centric data such as force fields and
obstacle/landmark positions into these modalities would give the
human operator a better understanding of the state of a robot swarm
in the field. Brooks changed the course of artificial intelligence
(AI) by arguing that the world is its own best model
\cite{Brooks1990}. In a similar fashion, we now argue that the
best-placed entity to obtain feedback about a swarm’s ‘body’ is the
operator’s body.

\begin{table*}[htb]
\centering
\begin{tabular}{|p{2.5cm}|p{3cm}|p{3.5cm}|p{3.5cm}|p{3.5cm}|}
\hline
\begin{center} \textbf{Device} \end{center} & \begin{center} \textbf{Example of current use} \end{center} & \begin{center} \textbf{Benefits} \end{center} & \begin{center} \textbf{Limitations} \end{center} & \begin{center} \textbf{Mitigation strategy} \end{center}\\
\hline
Wearable gesture recognition device (e.g., Myo) & \begin{itemize} \item Gaming apps \item Drone control \item Virtual reality \end{itemize} & \begin{itemize} \item Wearable gesture recognition \item Easy setup and use \item High CCR \item Solid SDK \end{itemize} & \begin{itemize} \item Limited range of gestures \end{itemize} & \begin{itemize} \item Custom profiles \end{itemize}\\
\hline
Portable deictic-recognition device (e.g., Leap Motion) & \begin{itemize} \item Gaming apps \item Virtual reality \item Academic research \end{itemize} & \begin{itemize} \item Precision joint tracking \item Small size \item Spacious interaction space \end{itemize} & \begin{itemize} \item USB connection required \item Support frame required for wearable applications \end{itemize} & \begin{itemize} \item Raspberry Pi SBC \item 3D Printed frame \end{itemize}\\
\hline
Wearable advanced haptic device (e.g., 3\(^{\text{rd}}\) Space, KOR-FX) & \begin{itemize} \item Gaming apps \end{itemize} & \begin{itemize} \item Frontal and back haptic interaction \item Solid SDK \end{itemize} & \begin{itemize} \item USB connection required \item Air compressor required (3\(^{\text{rd}}\) space) \item Sound connection required (KOR-FX) \end{itemize} & \begin{itemize} \item Raspberry Pi SBC \item Li-ion rechargeable battery \item Small pocket to carry air compressor \end{itemize}\\
\hline
\end{tabular}
\caption{\label{tab:orgtable1}
Features of all key components of the proposed interface.}

\end{table*}

Table \ref{tab:orgtable1} outlines the main benefits, limitations, and
mitigation strategies for all devices proposed before. First, Myo
provides wireless gesture recognition without complex infrastructure
settings. However, its gesture range is limited to the ones depicted
in Fig. \ref{fig:orgparagraph12}. Even though this does not represent a
problem at the current, proof–of-concept stage, advanced users could
extend the repertoire of recognizable gestures as their applications
require. In that case, custom gestures could be played, recorded,
and stored by using the device’s built-in EMG monitoring tool.

Second, Leap Motion offers high precision joint tracking for both
hands. Even though Leap Motion has a small size and large
interaction space, it requires a USB connection to operate. The
inclusion of a Single Board Computer (SBC) such as Raspberry
Pi\footnote{\url{http://www.raspberrypi.org/products/raspberry-pi-3-model-b/}} in the interface configuration could solve this
problem, as well as provide additional computing and communication
capabilities to the whole interface. In addition, the Leap Motion
sensor needs to be located such that it can clearly sense the
operator’s hands. To this end, a support frame could be attached to
the haptic vest to place it in a suitable location. 3D-printing this
support frame would ensure precise conformance with the vest’s
dimensions.

Third, the 3\(^{\text{rd}}\) Space gaming vest provides a large haptic interaction
body area (both frontal and back parts) and a solid software
development kit (SDK) with which to build applications. However,
device operation requires a USB connection as well as a portable air
compressor. The former requirement could be solved by the inclusion of
a mini PC (in the same way as described before) in the wearable
interface; the latter could be solved by the inclusion of a small
pocket in the vest to carry the air compressor, as well as a Li-ion
rechargeable battery that could power the whole system. 

In a nutshell, the combination of a wearable gesture recognition
device that can detect high-level intentions, a portable device that
can detect Cartesian information and finger movements, and a
wearable advanced haptic device that can provide real-time feedback
is a promising scheme for a general-purpose wearable interface for
HSI applications. As far as this author knows, this work is the
first to envision a wearable HSI interface that separates the input
and feedback components of the classical control loop (input,
output, feedback). Moreover, the proposed interface is suitable for
both indoor and outdoor environments. In addition, such an enhanced
interface might be able to provide other advanced interaction and
interesting control capabilities: some are described below.

\subsection{Enhanced haptic feedback}
\label{sec:orgheadline4}

Previous swarm robotics research involving haptic feedback has only
explored the use of traditional static devices normally attached to
a desktop computer terminal
\cite{Setter2015,Setter2015-leaderfollower}. This work is the first to
suggest the use of wearable technology in a wider range of
scenarios to allow the operator to obtain richer feedback. For
instance, classical force feedback could be conveyed to the
operator using this wearable technology and, therefore, increase
the immersion experienced by the operator. Moreover, the operator
might obtain additional information using this technology by
utilizing different pulse or vibration patterns (e.g.,
heartbeat-like pulse patterns could serve to communicate the
battery status of swarm robots). Finally, by decoupling the
feedback and input components of the interface, a more robust and
fault-tolerant interface can be achieved.

\subsection{Timing based input}
\label{sec:orgheadline5}

Recent research \cite{Nagavalli2014,Nagavalli2015} has demonstrates
that improper timing of control input by operators can lead to
problems when commanding a swarm of robots, such as group
fragmentation (i.e. unintentional division of the swarm). Group
fragmentation causes delays in coordination, as well as motion and
sensing errors that hurt the performance of swarm tasks. Operators
who issue commands frequently showed higher levels of swarm
fragmentation than those who allowed the swarm to adjust between
new commands. Optimal timing studies are just beginning to emerge,
and they are an interesting area for future research.

An interface that could determine optimal human timing as well as
provide guidance and assistance could be crucial to achieving
effective interaction between human operators and robot swarms. The
proposed interface outlined in this work offers a suitable platform
to conduct research on optimal human timing algorithms since it has
all the necessary components to develop effective models. First, an
embedded computing unit (e.g., Raspberry Pi) gathers input data
from the robotic swarm and calculates proper timing threshold
parameters. Second, an advanced haptic component (e.g., 3\(^{\text{rd}}\)
Space Gaming Vest or KOR-FX) provides intuitive patterns based on
previously calculated timing parameters to the operator to assist
with his/her decision-making. Finally, high-level gesture (e.g.,
Myo) and deictic recognition (e.g., Leap Motion) components send
commands to the robot swarm after the feedback is taken into
account.

\subsection{Hierarchical control}
\label{sec:orgheadline6}

Several recent surveys \cite{Kolling2015,Brambilla2013} have pointed
out that one of the main problems in the swarm robotics field is
that robotic swarms cannot switch between different behaviors
during the same mission at present.

Due to the loosely-coupled settings of an interface composed of
different wearable parts, it may be possible to create a taxonomy
of commands suitable for a wide range of robot behaviors. For
instance, high-level commands such as gestures could serve as a
switch mechanism between different robot behaviors, while deictic
movements could command the swarm within that specific mode of
operation 

\subsection{Simple interface}
\label{sec:orgheadline7}

Early studies \cite{Nunnally2012,Walker2012} in the field of swarm
visualization and representation indicated that simplifying the
large state of a swarm to a lower-dimensional representation can be
beneficial when controlling a group of robots. Reducing the amount
of noise as well as fusing information to simplify the problem of
determining a swarm’s state are further promising topics in the HSI
field.

The proposed interface outlined in this paper allows the
possibility of mapping the state of the robot swarm to an
operator’s body through haptic feedback. This capability could
dramatically increase the amount of status information available
without increasing the complexity of its representation.

\section{Conclusions}
\label{sec:orgheadline9}

Robotic swarms are expected to become an integral part of emerging
technologies and open the door to future economic possibilities.
However, the lack of a general purpose human-swarm interface that
provides seamless interaction between a human operator and a group of
robots confines the field to academic laboratories. 

Recent advances in gaming technology have brought sophisticated
devices that, if combined, could further advance the HSI discipline.
Wearable gesture recognition devices recognize and interpret
gestures in a wide range of scenarios. In addition, they provide the
mobility required for an operator to command a group of robots in
both remote and proximity interaction scenarios. Also, new haptic
devices such as gaming vests provide a means for an operator to
receive haptic feedback without interfering with his/her input
signal. At the same time, they constitute a novel platform to obtain
rich feedback without increasing the complexity of the swarm’s
status information.

The aim of this work is to incorporate the underlying principles of
these two novel technologies into a general-purpose HSI interface
that is able to control adaptive robotic swarms, which can be
controlled in a natural and seamless manner by human operators in
order to tackle complex tasks. Potential applications, such as
remote and proximity interaction with swarms of unmanned aerial
vehicles (UAVs), could be achieved without complex calibration and
infrastructure settings. Finally, outdoor applications (e.g.
agricultural tasks) could greatly benefit from the proposed
interface.

\bibliographystyle{plain}
\bibliography{References}
\end{document}